# The limit of artificial intelligence: Can machines be rational?

Tshilidzi Marwala

University of Johannesburg

South Africa

Email: tmarwala@gmail.com

**Abstract**

This paper studies the question on whether machines can be rational. It observes the existing reasons why humans are not rational which is due to imperfect and limited information, limited and inconsistent processing power through the brain and the inability to optimize decisions and achieve maximum utility. It studies whether these limitations of humans are transferred to the limitations of machines. The conclusion reached is that even though machines are not rational advances in technological developments make these machines more rational. It also concludes that machines can be more rational than humans.

**Introduction**

One of the most interesting concepts invented by humans is rationality (Anand, 1993; Marwala, 2014&2015). According to the google dictionary rationality is "the quality of being based on or in accordance with reason or logic." It is quite ironic that humans, who are by all accounts, irrational agents would invent the concept of rationality. Humans have also invented intelligent machines. These machines are used to predict the weather, to buy and sell stocks in the market, for face recognition etc. Given the growing importance of these machines, should they be designed to be rational? Or rather more controversially, is it technologically possible to design them to be rational? Given the fact that they are interacting with human beings, is it useful to design them to be rational. Rational machines maximize their utility. Is it possible for these machines to maximize their utility? Will we ever know that the maximum expected utility is indeed the ultimate maximum or in mathematical nomenclature a global maximum. In mathematics there are two types of optimization problems (in the context of this paper for the identification of the maximum utility) and these are convex and non-convex problems (Borwein and Lewis, 2000). Convex problems are problems where we can know for sure that the maximum utility that we have identified is the real deal. The non-convex problems are problems which we can never know that the maximum utility we have identified is the real deal. In turns out that practical and complex problems are non-convex and, therefore, we cannot be able to know if we have maximized utility.

**What is rationality?**

So what is this elusive concept called rationality. Rationality is the art of making logical decisions. It manifests itself by maximizing utility. For example, a businessman selling tomatoes maximizes utility by maximizing profit. A stock broker maximizes utility by maximizing the amount of money that they attain from playing in the stock market. What is then the mechanism of maximizing utility? There are two elements of maximizing utility and the first is to use information and the second is to process that information. For us humans on seeking to make rational decisions, we gather information through our five senses i.e. hearing, seeing, smelling, touching and tasting. Once we have the information, we make sense of such information using our biological computer which is our brain. The way a human senses information is limited. For example, if I give a person data with 10000 variables there is no way any of our five senses will be able to make sense of these variables. Our biological brain is also limited in many ways primarily because it did not evolve to analyze 10000 variables. The species that survived in the evolutionary trajectory did not survive because of knowing how to analyze 10000 variables. Consequently, humans are hopelessly irrational agents.

**What did Herbert Simon say?**

Nobel laureate Herbert Simon realizing this limitation of humans introduced the concept called bounded (i.e. limited) rationality (Simon, 1982). He stated that given the fact that humans can never have all the information they need to make a decision and that their biological computer i.e. the brain is not a perfect machine by a stretch of any imagination and that we can never know if we have identified the maximum utility, human beings are, therefore, condemned to making at best bounded rational decisions. He then went on further and stated that such decision making is merely a satisficing activity. He derived satisficing by hybridizing the terms satisfying and sufficing. Herbert Simon never prescribed where these bounds that limit rationality are located in the decision space but merely specified that these bounds exist. Daniel Kahneman and Amos Tversky did ground breaking work of trying to identify the limits of these bounds and the best they could come up with was that humans do not maximize utility but minimize losses (Kahneman, 2011). In simple terms humans do not make decisions to maximize but to avoid risk. This simply means that humans are risk averse agents. Of course this conclusion can be independently reached by using the theory of evolution where species that survived were the ones that avoided risks rather than those that sought risks. Kahneman and Tversky called this risk averse character of human *prospect theory*.

**What happens when we replace humans with machines?**

If humans are fundamentally limited by prospect theory, what happens when we replace them with machines? Are machines limited by bounded rationality? Are they as risk averse as we humans are? Machines are what we make them to be. If we want to make machines risk averse, then they become risk averse. But can we make them rational and thus maximize utility? Is it possible for humans who cannot maximize utility to create machines that maximize utility? One way of building intelligent machines is by using artificial intelligence (AI) (Nelwamondo and Marwala, 2006). Artificial intelligence is the art of building intelligent machines. AI has been successful on understanding HIV (Marwala, 2009), on building computers that can model complex mechanical structures (Marwala, 2010), on predicting interstate conflict (Marwala and Lagazio, 2011), on identifying faults in complex structures (Marwala, 2012), and on understanding the financial markets (Marwala, 2013; Marwala and Hurwitz, 2017). The way AI works is that it takes input variable *x* to give an output, *y,* and this is mathematically written as follows: $y=f(x)$. Here f is the AI model and much details on this can be found in Marwala (2018). But AI does not remove the limitation of imperfect and incomplete data *x* and relies on the model f which is never perfect according to the statistician George Box (Box, 1976). There is another limitation, which is that the processing of the information by machines using computers is not ideal because of the limitation of digital computing. Therefore, AI with all its successes is rationally limited!

**Digital and Quantum Computing**

Conventional computer machines analyze information in digital format. Digital format means all the information is represented as ones and zeros. The processing power of a digital computer has been doubling every year and this phenomenon is called the Moore's Law (Moore, 1965). This means decisions which are executed using a digital computer this year are less efficient than the same decisions if executed next year. This means decisions are more rationally executed in the future than now. This means the bounds of rationality are shifted for better rationality if executed in the future as compared to if they are executed now. This is the case until digital computers reach Moore's Law's physical limits. Quantum computing is another paradigm that has emerged over the last few years (Feynman, 1982). Quantum computing is based on two concepts and these are quantum superposition and entanglement. Entanglement is what Einstein called "spooky action" because it entails instantaneous communication. While digital computing uses only two states which are ones and zeros, quantum computing uses these two states and the superposition of these two states. Quantum computers are more efficient than digital computers. When digital computers reach their physical limits (the end of Moore's Law) then quantum computers will take over and perhaps usher us into the quantum Moore's Law. In other words, the bounds of rationality will be improved in the quantum domain until another physical limit (quantum Moore's Law) is reached. Ultimately, there has to be a computational limit by which no further computational gains can be obtained. This computational limit remains an open question that requires further research.

**All models are wrong**

Biological computers have limitations whereas AI models have their own limitations. George Box made a far reaching observation by stating that all models are wrong. He argued that they are wrong for the simple reason that they attempt to recreate reality and, therefore, are not the real deal. In Physics we build models that try to recreate reality. These models are built by mathematically attempting to create reality. Such models include the Einstein's theory of general relativity or Max Planck's quantum theory (Planck, 1900; Einstein, 1915). These models are physically realistic because they attempt to model the physical reality from the understanding of the observed mechanisms. AI models are not physically realistic. They take observed data and fit sophisticated yet physically unrealistic models. Because of this reality they are black boxes with no direct physical meaning. Because of this reason they are definitely wrong yet are useful. If in our AI equation $y=f(x)$, f the model is wrong as stipulated by Box then despite the fact that this model is useful in the sense that it can reproduce reality, it is wrong. Because it is wrong, when used to make a decision such a decision cannot possibly be rational. A wrong premise cannot result in a rational outcome! In fact the more wrong the model is the less rational the decision is.

**Can machines be rational?**

Now coming to the core of this paper, is it possible to design a rational machine? Firstly, a rational machine should use all the available information. Theoretically, one can design an autonomous and intelligent software agent that goes to the internet and search for all relevant information and uses this to make a decision. If that can be achieved, then this machine can use AI to process the gathered information and make a rational decision. Unfortunately, there is a trade-off between the amount of information used and the accuracy of the AI model, a situation akin to the Heisenberg uncertainty principle (Heisenberg, 1927). However, for this decision to be rational the whole process of gathering and processing information will have to be optimized for maximum utility. With the exception of few convex problems, we can never know whether the decision is globally optimized or not. The expression that "all models are wrong" will render the whole process wrong in any case. So even though AI machines can be logical, they cannot be rational, thereby, making the theory of bounded rationality a fundamental theory of decision making. The best that can be done to the theory of bounded rationality is to observe that technology helps to improve the bounds of rational decision making until some physical limit is reached, an observation that Marwala (2014&2015) calls *flexibly-bounded rationality*. So in conclusion, machines can be logical but never fully rational.

**Conclusion**

This paper studied the question of whether machines can be rational. It examined the limitations of machine decision making and these were identified as the lack of complete and perfect information, the imperfection of the models as well as the inability to identify the global optimum utility. This paper concludes that machines can never be fully rational and that the best they can achieve is to be bounded rationally. However, machines can be more rational than humans.